\documentclass[letterpaper,10pt,conference]{ieeeconf}
\IEEEoverridecommandlockouts
\usepackage{algorithm}
\usepackage{algorithmic}
\usepackage{amsfonts}
\usepackage{amsmath}
\usepackage{amssymb}
\usepackage{array}
\usepackage{bm}
\usepackage{breakcites}
\usepackage{caption}
\usepackage{color}
\usepackage{comment}
\usepackage{float}
\usepackage{gensymb}
\usepackage{graphicx}
\usepackage{multirow}
\usepackage{pdfpages}
\usepackage{pgfplots}
\usepackage[caption=false,font=footnotesize,subrefformat=parens,labelformat=parens]{subfig}
\usepackage{textcomp}
\usepackage{tikz}
\usepackage{pgfplots}
\usepackage{wrapfig}
\usepackage{xcolor}
\usepackage{tabularx}
\usepackage{tikz}

\usetikzlibrary{spy}

\PassOptionsToPackage{hyphens}{url}\usepackage[breaklinks=true,hidelinks=true]{hyperref}

\graphicspath{{figures/}}

\definecolor{dargreen}{rgb}{0.0, 0.5, 0.0}

\title{Evaluating Uncertainty Calibration for Open-Set Recognition}

\author{Zongyao Lyu, Nolan B. Gutierrez, and William J. Beksi
\thanks{The authors are with the Department of Computer Science and Engineering,
        University of Texas at Arlington, Arlington,
        TX, USA.
        Emails:
        zongyao.lyu@mavs.uta.edu,
        nolan.gutierrez@mavs.uta.edu,
        william.beksi@uta.edu
        }
}

\begin{document}
\maketitle
\pagestyle{plain}

\begin{abstract}
Despite achieving enormous success in predictive accuracy for visual
classification problems, deep neural networks (DNNs) suffer from providing
overconfident probabilities on out-of-distribution (OOD) data. Yet, accurate
uncertainty estimation is \textit{crucial} for safe and reliable robot autonomy.
In this paper, we evaluate popular calibration techniques for open-set
conditions in a way that is distinctly different from the conventional
evaluation of calibration methods on OOD data. Our results show that closed-set
DNN calibration approaches are \textit{much less effective} for \textit{open-set
recognition}, which highlights the need to develop new DNN calibration methods
to address this problem.
\end{abstract}

\section{Introduction}
\label{sec:introduction}
Deep neural networks (DNNs) have been very successful at vision-based tasks,
which has led to their widespread deployment for real-world applications (e.g.,
object recognition, image segmentation, etc.). Nevertheless, a challenge of
DNN-based vision systems lies in the network's inclination to produce
overconfident predictions during inference, especially when facing categories
not seen during training (i.e., out-of-distribution (OOD) data)
\cite{hein2019relu}. Within the realm of robotics, this motivates research
questions such as how much trust can we put into the predictions of a DNN when
misclassifications may have catastrophic consequences
\cite{sunderhauf2018limits}? Methods have been proposed to mitigate the
overconfidence problem by calibrating the predictive probabilities
\cite{guo2017calibration}, or estimating the predictive uncertainties
\cite{gal2016dropout, lakshminarayanan2017simple}, for various vision-based
objectives (e.g., image classification \cite{zhao2021calibrating}, semantic
segmentation \cite{kendall2017bayesian}, and object detection
\cite{kuppers2020multivariate,lyu2021uncertainty}). 

When deployed in real-world environments, we desire a DNN to have the
discriminative ability to separate query inputs into either \textbf{known}
(i.e., seen during training) or \textbf{unknown} (i.e., not yet seen during
training) classes. This problem has previously been formalized and studied under
the context of \textbf{open-set recognition} (OSR) \cite{scheirer2012toward}.
OSR is the extension of object recognition from closed-set to open-set
conditions, where classes outside of the training data can appear at inference
time. Specifically, OSR trains a model on $K$ classes of an $N$ class training
dataset. Then, at test time, the model is faced with $N$ different classes of
which $N-K$ were not seen during training. An explicit none-of-the-above (or
unknown) class placeholder (i.e., the $K+1$ class) is added during inference.
OSR aims to assign correct labels from $K$ known classes to seen test-time
samples, while detecting and placing unseen samples to the unknown class.
Robotic applications of OSR include the division of semantically known places
from semantically unknown \cite{sunderhauf2016place}, failure identification for
self-driving cars \cite{ramanagopal2018failing}, and entity detection
\cite{vieira2021open}. 

\textbf{Calibration} refers to the issue of obtaining a model that has a
predictive probability that reflects the true correctness likelihood.  More
formally, a model is \textit{well calibrated} if outcomes predicted to occur
with probability $p$  occur approximately $p$ fraction of the time
\cite{naeini2015obtaining}. In robotics, there is a clear need for uncertainty
calibration of policies (e.g., autonomous driving
\cite{nozarian2020uncertainty}), especially for OSR. Previous work has evaluated
calibration methods on both in distribution (ID) cases and under distributional
shifts \cite{ovadia2019can,minderer2021revisiting}. The experimental settings in
\cite{ovadia2019can,minderer2021revisiting} follow those in standard OOD
detection works where a model trained on the ID set is evaluated for its
reliability in identifying test images as either ID or OOD
\cite{hendrycks2016baseline}. Although dealing with similar problems, OOD
detection and OSR have been studied separately with different evaluation
protocols. To the best of our knowledge, the evaluation of calibration methods
for the OSR problem (i.e., \textbf{open-set calibration}) \textit{has not been
explored}. Hence, our work serves as a step toward this direction.

\section{Open-Set Recognition Evaluation}
\label{sec:open-set_recognition_evaluation}
The first step of our evaluation is to enhance a model to perform OSR. To do
this, we create a baseline by thresholding on the maximum probability which
indicates the mostly likely predicted class. We also apply OpenMax
\cite{bendale2016towards}, a well-known OSR method, for comparison. Following
the standard experimental setup in OSR works, we use ResNet \cite{he2016deep} as
the backbone classification network and evaluate on the CIFAR10 benchmark
dataset \cite{krizhevsky2009learning}. To test under OSR conditions, we follow
the most common data partitioning protocol \cite{neal2018open,
yoshihashi2019classification,oza2019c2ae,perera2020generative,zhang2020hybrid,
zhou2021learning}. Specifically, we split each dataset at random such that 6
classes are chosen to be known and the remaining 4 classes are unknown. We
repeat the experiment over 5 runs and report the average score. Thus, the model
is trained with 6 known classes and tested on all 10 classes. We map the
randomly selected 6 known classes using indices 0-5 and set the other 4 unknown
classes to start at index 6. A perfect OSR model should be able to assign
correct labels to images belonging to any of the 6 known classes and identify
images from all other classes as unknowns.

In the second step of our evaluation, we calibrate the trained model on the
validation set in order to improve calibration during inference.
Calibration is performed using \textbf{temperature scaling}
\cite{guo2017calibration}, a simple and effective technique that makes
adjustments to a model's logits by a single temperature parameter $T$.
\textbf{Logits} are probability vectors produced by a network before the softmax
layer for a given input image. Additionally, $T$ is optimized on a held-out
validation set and it can rescale a model's confidence without affecting
accuracy. After acquiring an OSR model, we apply temperature scaling on the
model's predictive confidence and evaluate the calibration performance in this
novel problem setting. Concretely, to apply temperature scaling we reserve 10\%
of the training data as a validation set for optimizing $T$ and use the
remaining 90\% for training the model. To evaluate model calibration, we use the
\textbf{expected calibration error} (ECE) \cite{naeini2015obtaining} to measure
the absolute difference between predictive confidence and accuracy. As a result
of its simplicity and effectiveness, ECE is the most commonly used metric for
model calibration. We also evaluate the \textbf{Brier score} within the
framework of OSR as follows. If a trained classifier's prediction $\hat{\bm{Y}}
\in \mathbb{R}^{N \times K},$ where $N$ is the number of samples and $K $ is the
number of known classes, then we can obtain a new prediction
$\hat{\bm{Y}}_{\textrm{OSR}} \in \mathbb{R}^{N \times (K + 1)}$ that takes into
account the unknown class. For each prediction $\hat{\bm{y}} \in \mathbb{R}^K$
we estimate the probability of being unknown, i.e.,
\begin{equation}
  \hat{y}_{\textrm{unknown}} = \prod\limits_{i = 1}^K (1 - \hat{y}_i).
\end{equation}
Next, we calculate the Brier score by
\begin{equation}
  \frac{1}{N}\sum\limits_{i = 1}^{N}\sum\limits_{j = 1}^{K}(\hat{Y}^{ij}_{\textrm{OSR}} - Y^{ij})^2,
\end{equation}
where the predicted probability and ground truth of the $j $th class are $
\hat{Y}^{ij}_{\textrm{OSR}}$ and $ Y^{ij}$, respectively. 



\section{Evaluation Results}
\label{sec:evaluation_results}
\begin{figure}
\centering
\subfloat{
  \includegraphics[width=.38\columnwidth]{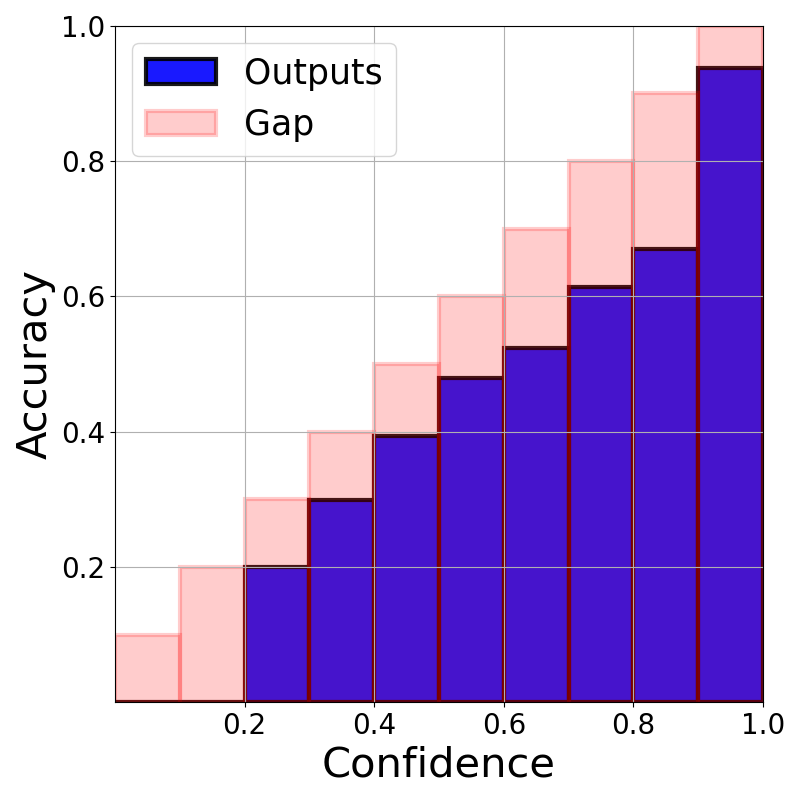}\hspace{1mm}}
\subfloat{
  \includegraphics[width=.38\columnwidth]{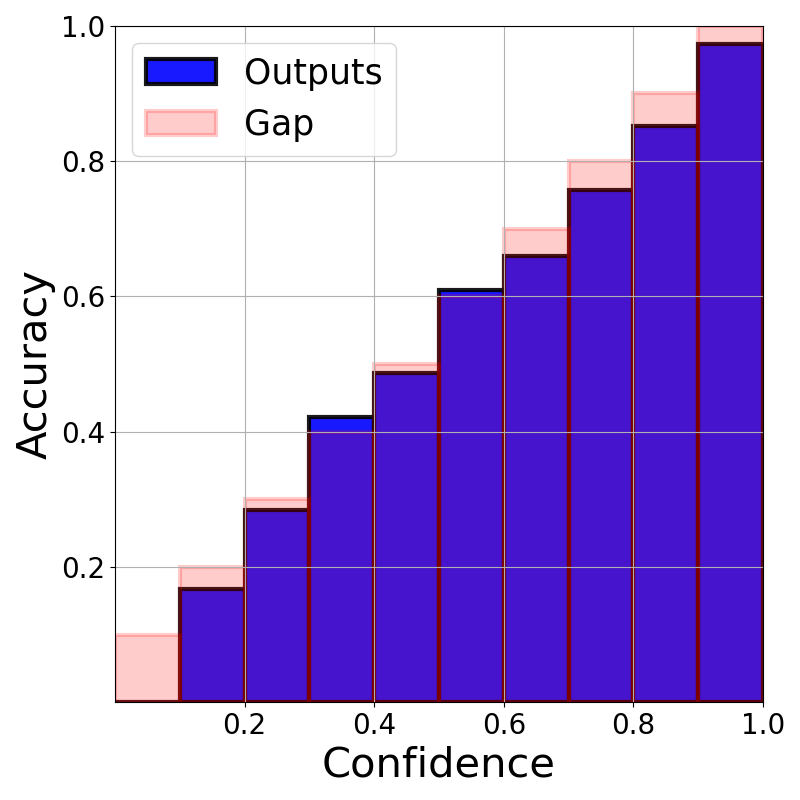}\hspace{1mm}}\\
\subfloat{
  \includegraphics[width=.38\columnwidth]{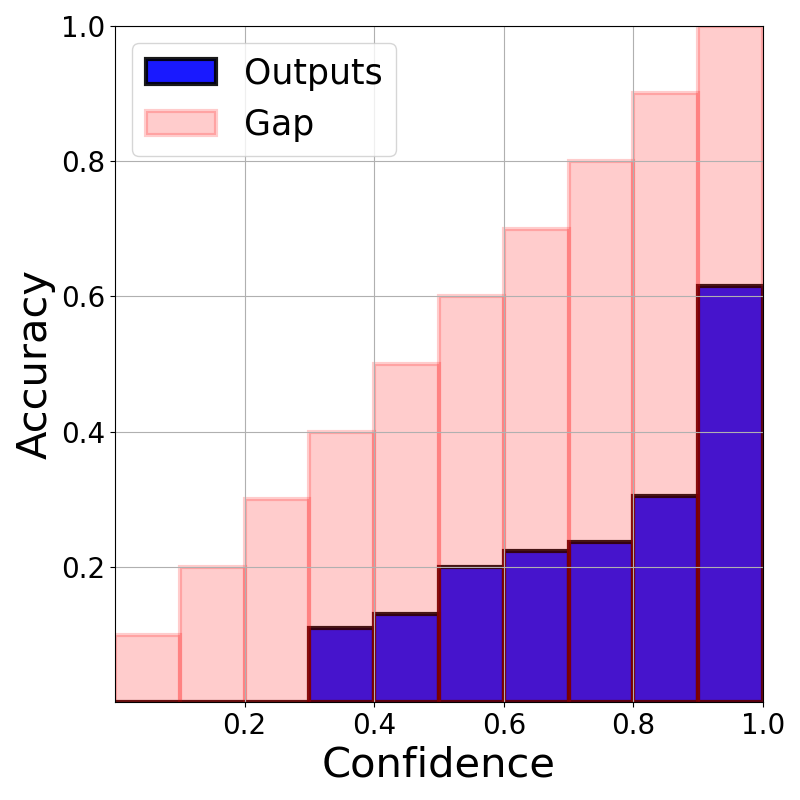}\hspace{1mm}}
\subfloat{
  \includegraphics[width=.38\columnwidth]{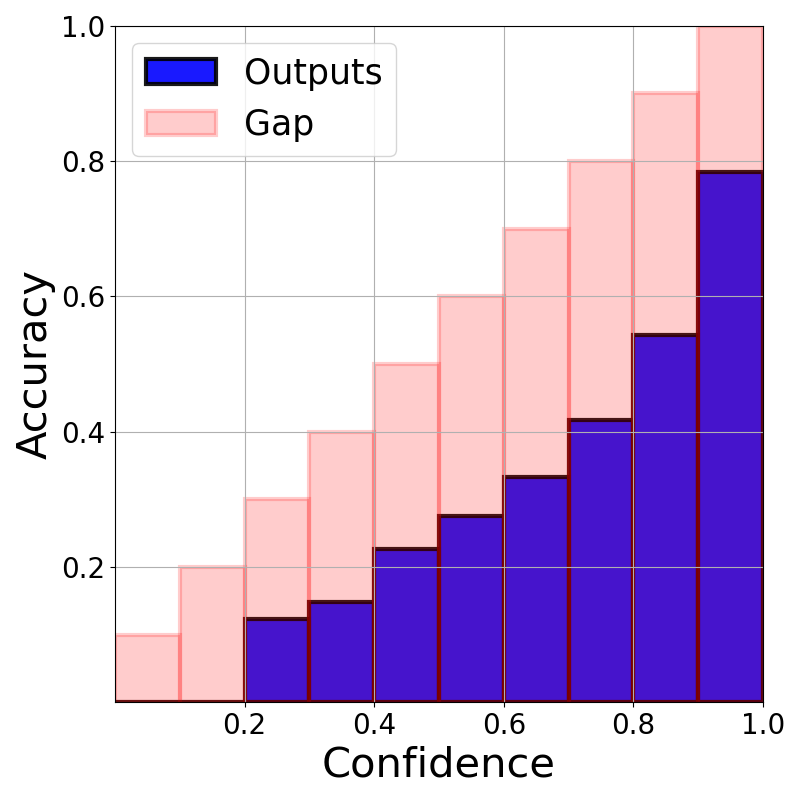}\hspace{1mm}}\\
\subfloat{
  \includegraphics[width=.38\columnwidth]{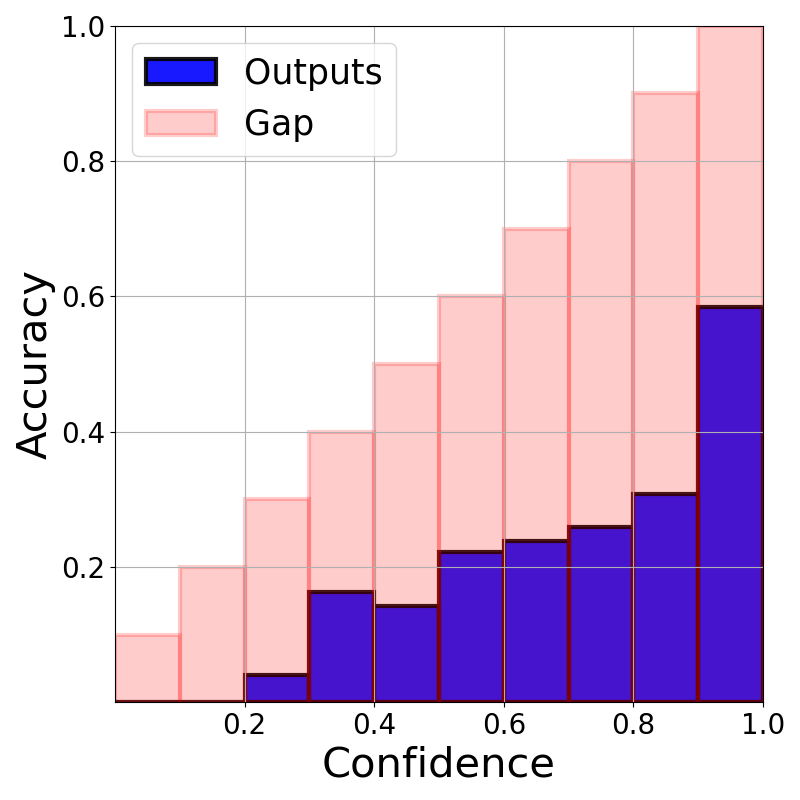}\hspace{1mm}}
\subfloat{
  \includegraphics[width=.38\columnwidth]{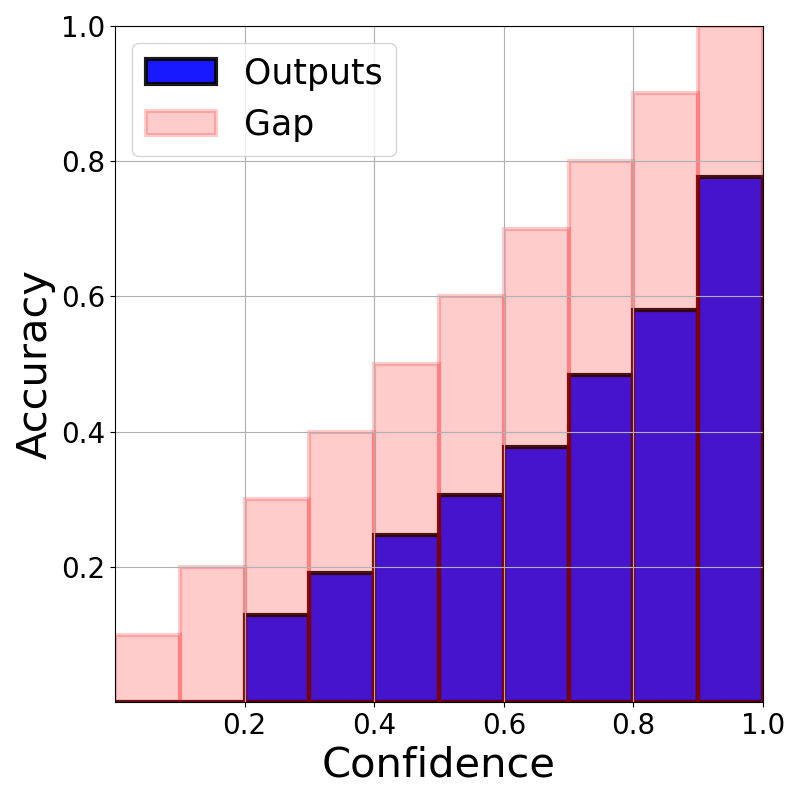}\hspace{1mm}}\\
\caption{The reliability diagrams of the \textit{closed-set} (top row),
\textit{open-set} (middle row), and \textit{open-set + OpenMax} (bottom row)
methods before (left column) and after (right column) performing temperature
scaling calibration.}
\label{fig:close_reliability_diagram}
\vspace{-2mm}
\end{figure}

We visualize model calibration via \textbf{reliability diagrams}
\cite{degroot1983comparison, niculescu2005predicting}, where the diagonal
represents perfect calibration and any deviation from a perfect diagonal
indicates miscalibration. The top rows of
Fig.~\ref{fig:close_reliability_diagram} and Table~\ref{tab:resnet_cifar10}
present the reliability diagrams and scores for closed-set classification before
and after calibration by temperature scaling. We see that temperature scaling
significantly improves the model's ability to match the true correctness
likelihood in calibrating the predictions for this conventional 10-class
closed-set classification problem. In the middle rows of
Fig.~\ref{fig:close_reliability_diagram} and Table~\ref{tab:resnet_cifar10}, the
reliability diagrams and scores of a classifier operating in open-set conditions
by thresholding on its predictive probabilities before and after calibration by
temperature scaling are shown. In this scenario, temperature scaling still
provides a clear improvement over uncalibrated models. Nonetheless, there exist
clear gaps between temperature scaling calibrated confidence and perfectly
calibrated confidence thus indicating that the calibration problem for open-set
conditions is much more challenging.
The bottom rows of Fig.~\ref{fig:close_reliability_diagram} and
Table~\ref{tab:resnet_cifar10} demonstrate the reliability diagrams and scores
of a classifier operating in open-set conditions by applying OpenMax before and
after calibration.  Although OpenMax improves classification accuracy, it
exhibits no obvious improvement in terms of calibration. In addition, we also
conducted these experiments with disparate classification networks such as
DenseNet \cite{huang2017densely} and EfficientNet \cite{tan2019efficientnet} on
the SVHN dataset \cite{netzer2011reading} (see appendix). In all of these
experiments, we observed the same findings as above.


\section{Conclusion}
\label{sec:conclusion}
In this paper, we assessed the efficacy of calibration methods for OSR. From the
evaluation results, we observe that the calibration problem is much more
challenging under open-set conditions in comparison to closed-set conditions.
Previous techniques that calibrate confidence really well for closed-set
classification only provide limited calibration performance for OSR. Moreover,
traditional OSR approaches improve accuracy, yet they do not take calibration
into consideration. Given these observations, we want to bring the community's
attention not only to the pursuit of more accurate OSR methods, but also
well-calibrated OSR systems that can perform reliably and safely in open-set
settings.


\begin{table}[H]
\centering
\begin{center}
\begin{tabular}{|c|c|c|c|c|}
\hline
{\bf Method} & {\bf Measure} & {\bf Brier} $\downarrow$ & {\bf ECE} $\downarrow$ & {\bf Accuracy} \\
\hline
\multirow{2}{*}{Closed-set} & Before calibration & 0.255 & 0.069 & \multirow{2}{*}{0.832}\\
& After calibration & 0.244 & 0.011 & \\
\hline
\multirow{2}{*}{Open-set} & Before calibration  & 0.814 & 0.346 & \multirow{2}{*}{0.603}\\
& After calibration & 0.647  & 0.230 &  \\
\hline
\multirow{2}{*}{\shortstack{Open-set + \\OpenMax}} & Before calibration  & 0.841 & 0.335 & \multirow{2}{*}{0.632} \\
& After calibration & 0.703  & 0.197 & \\
\hline
\end{tabular}
\end{center}
\caption{Brier and ECE scores for closed-set and open-set calibration on {\bf CIFAR10} using {\bf ResNet}.}
\label{tab:resnet_cifar10}
\end{table}



\bibliographystyle{IEEEtran}
\bibliography{IEEEabrv,evaluating_uncertainty_calibration_for_open-set_recognition}

\appendix

\begin{table}[H]
\centering
\begin{center}
\begin{tabular}{|c|c|c|c|c|}
\hline
{\bf Method} & {\bf Measure} & {\bf Brier} $\downarrow$ & {\bf ECE} $\downarrow$ & {\bf Accuracy} \\
\hline
\multirow{2}{*}{Closed-set} & Before calibration & 0.085 & 0.009 & \multirow{2}{*}{0.943}\\
&After calibration & 0.084 & 0.006 & \\
\hline
\multirow{2}{*}{Open-set} & Before calibration & 0.434 & 0.220 & \multirow{2}{*}{0.793}\\
&After calibration & 0.406 & 0.204 & \\
\hline
\multirow{2}{*}{\shortstack{Open-set + \\OpenMax}} & Before calibration & 0.496 & 0.224 & \multirow{2}{*}{0.811} \\
&After calibration & 0.489 & 0.218 & \\
\hline
\end{tabular}
\end{center}
\caption{Brier and ECE scores for closed-set and open-set calibration on {\bf SVHN} using {\bf ResNet}.}
\label{tab:resnet_svhn}
\vspace{-3mm}
\end{table}

\begin{table}[H]
\centering
\begin{center}
\begin{tabular}{|c|c|c|c|c|}
\hline
{\bf Method} & {\bf Measure} & {\bf Brier} $\downarrow$ & {\bf ECE} $\downarrow$ & {\bf Accuracy} \\
\hline
\multirow{2}{*}{Closed-set} & Before calibration & 0.246 & 0.070 & \multirow{2}{*}{0.835}\\
&After calibration & 0.234 & 0.012 & \\
\hline
\multirow{2}{*}{Open-set} & Before calibration & 0.788 & 0.357 & \multirow{2}{*}{0.645}\\
&After calibration & 0.643 & 0.269 & \\
\hline
\multirow{2}{*}{\shortstack{Open-set + \\OpenMax}} & Before calibration & 0.861 & 0.369 & \multirow{2}{*}{0.666} \\
&After calibration & 0.755 & 0.268 & \\
\hline
\end{tabular}
\end{center}
\caption{Brier and ECE scores for closed-set and open-set calibration on {\bf CIFAR10} using {\bf DenseNet}.}
\label{tab:densenet_cifar10}
\vspace{-3mm}
\end{table}

\begin{table}[H]
\centering
\begin{center}
\begin{tabular}{|c|c|c|c|c|}
\hline
{\bf Method} & {\bf Measure} & {\bf Brier} $\downarrow$ & {\bf ECE} $\downarrow$ & {\bf Accuracy} \\
\hline
\multirow{2}{*}{Closed-set} & Before calibration & 0.091 & 0.008 & \multirow{2}{*}{0.939}\\
&After calibration & 0.090 & 0.006 & \\
\hline
\multirow{2}{*}{Open-set} & Before calibration & 0.447 & 0.229 & \multirow{2}{*}{0.816}\\
&After calibration & 0.376 & 0.178 & \\
\hline
\multirow{2}{*}{\shortstack{Open-set + \\OpenMax}} & Before calibration & 0.507 & 0.231 & \multirow{2}{*}{0.822} \\
&After calibration & 0.483 & 0.209 & \\
\hline
\end{tabular}
\end{center}
\caption{Brier and ECE scores for closed-set and open-set calibration on {\bf SVHN} using {\bf DenseNet}.}
\label{tab:densenet_svhn}
\vspace{-3mm}
\end{table}

\begin{table}[H]
\centering
\begin{center}
\begin{tabular}{|c|c|c|c|c|}
\hline
{\bf Method} & {\bf Measure} & {\bf Brier} $\downarrow$ & {\bf ECE} $\downarrow$ & {\bf Accuracy} \\
\hline
\multirow{2}{*}{Closed-set} & Before calibration & 0.233 & 0.032 & \multirow{2}{*}{0.839}\\
&After calibration & 0.230 & 0.012 & \\
\hline
\multirow{2}{*}{Open-set} & Before calibration & 0.767 & 0.373 & \multirow{2}{*}{0.633}\\
&After calibration & 0.665 & 0.275 & \\
\hline
\multirow{2}{*}{\shortstack{Open-set + \\OpenMax}} & Before calibration & 0.841 & 0.334 & \multirow{2}{*}{0.636} \\
&After calibration & 0.719 & 0.151 & \\
\hline
\end{tabular}
\end{center}
\caption{Brier and ECE scores for closed-set and open-set calibration on {\bf CIFAR10} using {\bf EfficientNet}.}
\label{tab:efficientnet_cifar10}
\vspace{-3mm}
\end{table}

\begin{table}[H]
\centering
\begin{center}
\begin{tabular}{|c|c|c|c|c|}
\hline
{\bf Method} & {\bf Measure} & {\bf Brier} $\downarrow$ & {\bf ECE} $\downarrow$ & {\bf Accuracy} \\
\hline
\multirow{2}{*}{Closed-set} & Before calibration & 0.101 & 0.010 & \multirow{2}{*}{0.932}\\
&After calibration & 0.100 & 0.004 & \\
\hline
\multirow{2}{*}{Open-set} & Before calibration & 0.394 & 0.207 & \multirow{2}{*}{0.845}\\
&After calibration & 0.377 & 0.197 & \\
\hline
\multirow{2}{*}{\shortstack{Open-set + \\OpenMax}} & Before calibration & 0.466 & 0.209 & \multirow{2}{*}{0.857} \\
&After calibration & 0.450 & 0.193 & \\
\hline
\end{tabular}
\end{center}
\caption{Brier and ECE scores for closed-set and open-set calibration on {\bf SVHN} using {\bf EfficientNet}.}
\label{tab:efficientnet_svhn}
\vspace{-3mm}
\end{table}


\end{document}